\documentclass[pdflatex,sn-mathphys-ay]{sn-jnl}

\usepackage{graphicx}%
\usepackage{multirow}%
\usepackage{amsmath,amssymb,amsfonts}%
\usepackage{amsthm}%
\usepackage{mathrsfs}%
\usepackage[title]{appendix}%
\usepackage{textgreek} 

\usepackage{xcolor}%
\usepackage{textcomp}%
\usepackage{manyfoot}%
\usepackage{booktabs}%
\usepackage{algorithm}%
\usepackage{algorithmicx}%
\usepackage{algpseudocode}%
\usepackage{listings}%
\usepackage{pgf-pie}
\usepackage{adjustbox}
\usepackage{pgfplots}
\pgfplotsset{compat=1.18}
\usepackage{pgfplotstable}
\usepackage{filecontents}
\usepackage{pgfplots}
\usepackage{caption}
\usepackage{url}
\pgfplotsset{compat=1.18}

\theoremstyle{thmstyleone}%
\theoremstyle{thmstyletwo}%

\theoremstyle{thmstylethree}%

\begin{document}



\title[Article Title]{Overcoming Low-Resource Barriers in Tulu: Neural Models and Corpus Creation for Offensive Language Identification}


\author[1]{\fnm{Anusha } \sur{M D}}\email{anugowda251@gmail.com}

\author[1]{\fnm{Deepthi} \sur{Vikram}}\email{bangeradeepthi@gmail.com}

\author[2]{\fnm{Bharathi } \sur{Raja Chakravarthi}}\email{bharathi.raja@universityofgalway.ie}

\author*[1]{\fnm{Parameshwar} \sur{R Hegde}}\email{param1000@yahoo.com}

\affil*[1]{\orgdiv{Department of Computer Science}, \orgname{Yenepoya (Deemed to be University)}, \orgaddress{\street{Balmata}, \city{Mangalore}, \postcode{575002}, \state{Karnataka}, \country{India}}}

\affil[2]{\orgdiv{School of Computer Science},\orgname{ University of Galway} ,\orgaddress{\street{Galway}
  \country{Ireland}}
}


\abstract{Tulu, a low-resource Dravidian language predominantly spoken in southern India, has limited computational resources despite its growing digital presence. This study presents the first benchmark dataset for Offensive Language Identification (OLI) in code-mixed Tulu social media content, collected from YouTube comments across various domains. The dataset, annotated with high inter-annotator agreement (Krippendorff’s α = 0.984), includes 3,845 comments categorized into four classes: Not Offensive, Not Tulu, Offensive Untargeted, and Offensive Targeted. We evaluate a suite of deep learning models, including GRU, LSTM, BiGRU, BiLSTM, CNN, and attention-based variants, alongside transformer architectures (mBERT, XLM-RoBERTa). The BiGRU model with self-attention achieves the best performance with 82\% accuracy and a 0.81 macro F1-score. Transformer models underperform, highlighting the limitations of multilingual pretraining in code-mixed, under-resourced contexts. This work lays the foundation for further NLP research in Tulu and similar low-resource, code-mixed languages.}


\maketitle
\section{Introduction}\label{sec1}
Language is a vital tool for communication and enables people to express thoughts, emotions, and ideas. It is an important part of all human interactions and includes a system of written symbols and sounds \citep {sreelakshmi2024}. The linguistic landscape of India is extremely diverse, and hundreds of languages belong to various linguistic groups\citep{kulkarni2019linguistic}. Tulu \citep{bhat2019tulu}, spoken predominantly in the southern Karnataka regions and parts of Kerala, is among the Dravidian languages \citep{hegde2022corpus}. Tulu still lacks formal research and computational resources, although many speak of it and it has cultural importance \citep{durairaj2025}.

Tulu, a traditional oral language, is a less widely used native script that presents opportunities for the development of new linguistic resources \citep{bhat2019tulu}. 
In digital platforms, Tulu is most commonly written using either the Kannada or Latin alphabets, and it often appears in a code-mixed format blending elements from other languages. The linguistic characteristics of Tulu, combined with its informal and evolving nature, pose unique challenges for computational tasks \citep{hegde2022}. Although Tulu is predominantly spoken, its growing presence on digital platforms, particularly social media, has led to an increase in code-mixed text. This phenomenon is especially common in informal online communication, making the analysis of such texts both a challenge and an opportunity for linguistic research \citep{shetty2023poorvi}.

One of the critical areas in which computational linguistics can play a significant role is OLI. As online platforms continue to grow, the need for effective content moderation is becoming increasingly important. Offensive language, including hate speech, bullying, and toxic comments, is a growing concern on social media 
 \citep{chakravarthi2023offensive}. However, identifying such content is particularly challenging in under-resourced languages, such as Tulu, especially when the data are code-mixed. The lack of publicly available datasets and annotated corpora for Tulu OLI further complicates this task.

 As Tulu gains prominence in digital communication\citep{shetty2023poorvi} and the challenges associated with content moderation, this study aims to address the need for an OLI corpus in code-mixed Tulu texts.
This corpus was created from comments in Tulu language on YouTube. Comments span a wide variety of content types, such as movie trailers, short films, music, and trending videos.
The goal is to create an annotated dataset of offensive and non-offensive content in Tulu, which can then be used to train machine learning models to automatically detect offensive languages in such texts.
Contributions of the study:

\begin{itemize}
    \item The creation of a dataset for OLI in Tulu is presented, offering a valuable resource to advance research in this low-resource language. Additionally, this study establishes a benchmark for OLI in Tulu
    \item Deep learning models, including GRU, BiGRU, BiLSTM, LSTM,CNN BIGRU+SA 
    
    and BiLSTM with self attention (BiLSTM+SA), were used for offensive language classification. The Bidirectional GRU with Self-Attention(BiGRU+SA) demonstrated the highest performance, achieving an accuracy of 82\% and  code is available at \href{https://github.com/anushamdgowda/Codemixed-Tulu-OLI/tree/main}{Github}
    
    \item The study investigates the limitations of pre-trained transformer models, such as mBERT and XLM-RoBERTa, when applied to Tulu. These models performed suboptimally, emphasising the need for language-specific fine-tuning and adaptation for low-resource languages 
\end{itemize}

\section{Related works}\label{sec2}
OLI within code-mixed and under-resourced languages has emerged as a vital area of NLP research, notably given the proliferation of multilingual content on social media \citep{subramanian2022offensive}. Earlier investigations have scrutinized diverse modeling methods as well as datasets. Several early studies have explored modeling approaches and the creation of datasets for the detection of offensive or abusive language in code-mixed Indian languages \citep{suryawanshi2020multimodal, joshi2016towards}. 

A sentiment-annotated corpus comprising 15,744 Tamil-English code-mixed YouTube comments was introduced by \citep{chakravarthi2020corpus}. The comments were categorized into five sentiment classes. Their analysis revealed that classical models such as logistic regression and random forest outperformed deep learning models, likely due to the dataset's class imbalance. However, the study primarily focused on sentiment classification rather than identifying offensive language. The authors later expanded the dataset to over 60,000 comments, incorporating Tamil-English, Kannada-English, and Malayalam-English data, along with a multilingual sentiment analysis benchmark. Despite this expansion, the dataset did not specifically target offensive or abusive language categories.

Sentiment analysis and abusive language detection in Kannada-English code-mixed text has been explored, particularly in the work of  \cite{hande2020kancmd}. Their study found that while standard classifiers performed reasonably well for general categories, they struggled with ambiguous or subtle expressions, highlighting the complexities of identifying offensive content in code-mixed contexts. Likewise, Hegde et al. fashioned a trilingual sentiment dataset including Tulu, Kannada, and English. Within their observations pertaining to improving minority class detection, The issue of class imbalance and the need for clearer sentiment cues were highlighted. However, this effort did not specifically address the handling of offensive language.
In a related study, \cite{hegde2022corpus} developed a code-mixed sentiment dataset involving Tulu, Kannada, and English. Their research emphasized class imbalance and the importance of clear sentiment cues for improving minority class detection. They collected 7,171 YouTube comments in Tulu, Kannada, and English to create the first gold-standard code-mixed Tulu dataset for sentiment analysis. This dataset, annotated for various sentiment classes and code-mixing complexity, was used to train traditional classifiers such as Support Vector Machine (SVM), Multilayer Perceptron (MLP). The best F1-score of 0.60 was achieved by SVM and MLP, with 5-fold cross-validation showing 0.62 for SVM. However, the dataset was unbalanced, with the "Positive" class dominating and issues with minority class representation, especially "Negative" and "Not Tulu." Explicit sentiment cues helped improve classifier performance for these minority classes, making the dataset a valuable resource for future deep learning research and class balancing strategies in code-mixed sentiment analysis.

In a substantial contribution to Tulu NLP, Asha et al.\citep{asha2023kt2} employed deep learning models toward sentiment analysis of Tamil-Tulu YouTube comments, with this attaining an F1-score of 0.72. They also fashioned a Kannada-Tulu parallel corpus for machine translation via Transformer+LSTM+BPE architectures, coupled with this shows neural methods are feasible in low-resource settings. Their efforts failed to confront detrimental or objectionable material specifically. 

However,  \cite{garg2020annotated} centered on Hindi-English tweets that experts annotated regarding sarcasm, slang, plus sentiment, thereby assisting the strong multilingual datasets' development. Linguistic scope was augmented via their Hinglish work and also valuable perceptions furnished toward commercial and political discourse analysis. The endeavor stood as sentiment-centric in its form instead of being language-specific.

Despite the potential of transformer-based models like mBERT and XLM-Roberta for multilingual tasks, they operate with constrained effectiveness within low-resource and code-mixed scenarios. \cite{chakravarthi2022dravidiancodemix} employed Masked and Permuted Pre-training Network (MPNet)
 and CNN for OLI in Dravidian languages, yet the pretraining corpus's language representation greatly impacted performance. Although models like mBERT and XLM-RoBERTa have demonstrated effectiveness in multilingual contexts, their performance diminishes significantly for low-resource and code-mixed languages \citep{hu2020xtreme}.

Past investigations have established groundwork within sentiment analysis also multilingual processing regarding Indian languages,most studies either do not include Tulu or treat it as part of a broader multilingual dataset without dedicated analysis. Moreover, very few focus explicitly on offensive language, a gap this study seeks to address. The current work introduces the first manually annotated OLI corpus for code-mixed Tulu text and evaluates a range of traditional, deep learning, and transformer-based models to establish a benchmark for future research in this underrepresented language.

\section{Corpus Creation and Annotation}
\subsection{Data Collection and Filtering }
The quality of offensive language identification in Tulu was enhanced through a refinement step designed to preserve meaningful textual content.
While the dataset consists of 4,000 comments, this reflects the current availability of Tulu-language content on digital platforms. As an under-resourced language with a relatively recent emergence in online discourse, especially in informal and code-mixed contexts, the volume of publicly accessible data remains limited. The selected size balances data availability, annotation feasibility, and quality control, ensuring that the data set is representative and reliable for initial experimentation. This resource serves as a foundational benchmark for future expansion as Tulu’s digital presence continues to grow. The overall process of Tulu data creation is illustrated in Figure~\ref{figure3}.

\subsection{Annotation Setup}\label{sec3}
The annotation technique proposed by \cite{hegde2022corpus} was adopted to ensure consistency and cultural appropriateness in the Tulu offensive language dataset. A team of three annotators, consisting of two native Tulu speakers and one linguistic expert with advanced proficiency in the language, was engaged in this task.

Prior to annotation, formal orientation was undertaken by all annotators using Google Forms. The form provides comprehensive instructions, annotation category definitions, and sample examples to ensure mutual understanding of the task. All comments in the dataset were independently annotated by three annotators. All three annotators provided disparate labels, that is, no agreement; the comment was marked as ambiguous and excluded from the final dataset. The multistep annotation protocol was employed to ensure maximum consistency of the labels, reduce annotator bias, and maximize the overall reliability of the annotated dataset. As illustrated in Figure 2, the pipeline outlines the key steps involved in the creation of the dataset for Offensive Language Identification from YouTube comments.

\subsection{Inter-Annotator Agreement (IAA)}
To ensure the reliability of the annotations within the Tulu Offensive Language Identification dataset, an inter-annotator agreement analysis was conducted. Three annotators independently labeled each comment, and their level of agreement was measured using Krippendorff’s alpha a robust metric capable of handling multiple annotators and diverse data types \citep{zhao2018we}.

Krippendorff's alpha, calculated as 0.984 indicates an extremely high level of agreement between the annotators. The value is far above the minimum required of 0.80, indicating a high level of inter-rater reliability as proposed by \cite{krippendorff2011agreement}. A high level of agreement as such is an indicator of the reliability and efficacy of the annotation protocols, thus proving that the labeling process was replicable and reliable.
\subsection{Challenging Examples}
In the task of offensive language identification within code-mixed Tulu texts, certain comments exhibit ambiguous linguistic characteristics that complicate the annotation process \citep{hegde2022}. The presence of sarcasm, humor, and references to specific communities introduces challenges in deciding whether a comment is offensive or not \citep{chakravarthi2023offensive}. This section presents representative examples from the dataset that highlight these difficulties, demonstrating how the same text can be interpreted in multiple ways depending on tone and context.

\begin{enumerate}
\item {\color{red}Nikleg} {\color{red}posa} {\color{red}posa} {\color{blue}video} {\color{red}apload} {\color{red}manpare} {\color{red}danni?} {\color{red}Tulu} {\color{red}janagala} {\color{blue}comedy} \\
\textit{- Why are you uploading such useless videos? Tulu people's comedy }\\
The comment could be interpreted as humorous or disrespectful depending on tone. It references a group (Tulu people), so it’s difficult to decide if it’s general humor or targeted mockery.
\\
\item {\color{red}Latput} {\color{red}tulu} {\color{blue}movies} {\color{red}mukleg} {\color{red}ori} {\color{red}rowdi} {\color{blue}real} {\color{blue}comedy...} \\
\textit{-These silly Tulu movies are nothing—he's the real comedy rowdy.} \\
The comment may be humorous or mocking Tulu movies by contrasting them with real comedy. The sarcastic tone makes sentiment labeling challenging.
\\
\item {\color{blue}"Super} {\color{red}dada} {\color{blue}acting} {\color{blue}comedy} {\color{blue}scene} {\color{blue}bomb} {\color{red}ide"} \\
\textit{-Super brother, the acting in the comedy scene is a bomb} \\
This could be taken positively (a "bomb" as in amazing) or sarcastically (a "bomb" as in disaster), depending on tone/context.
\\
\item {\color{red}"Enchina} {\color{red}avaru} {\color{blue}comedy} {\color{blue}master"} \\
\textit{-What is this, he’s a comedy master? }\\
 Could be genuine praise or sarcastic, depending on how "comedy master" is delivered.

\end{enumerate}
According to the instructions given to the annotators, comments containing explicit offensive cues were used for annotation. However, some examples included subtle or implied offensive content that differed from what the explicit cues suggested. As a result, a few comments led to disagreements among annotators due to their nuanced nature.

\subsection{Dataset Statistics and Splits}
The final dataset consists of 3,845 user comments from Tulu YouTube videos, retained after applying the necessary filtering steps. Its size reflects the challenges inherent to resource development for low-resource languages like Tulu, particularly given the manual annotation process and the scarcity of high-quality digital text. 
\begin{figure}[!ht]
\centering
\begin{tikzpicture}
\pie[text=legend, radius=2, color={blue!40, orange!50, green!40}]
    {70/Train (2692), 15/Development (577), 15/Test (576)}
\end{tikzpicture}
\caption{Dataset Split for Tulu Offensive Language Dataset}
\label{fig:tulu_pie_chart}
\end{figure}
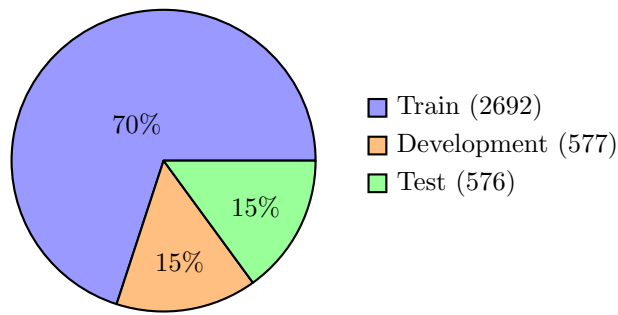
Figure~\ref{fig:tulu_trend} illustrates the rising interest in the Tulu language over recent years, based on Google Trends data. This upward trend indicates growing online engagement with Tulu, which helps justify the current dataset size and suggests that more data may become available in the near future. Despite its scale, the dataset offers a strong benchmark with high inter-annotator agreement and supports meaningful model evaluation, as demonstrated in this study.
The dataset statistics, including the number of comments, sentences, total words, and vocabulary size, are summarised in Table \ref{tab:label_distribution}.

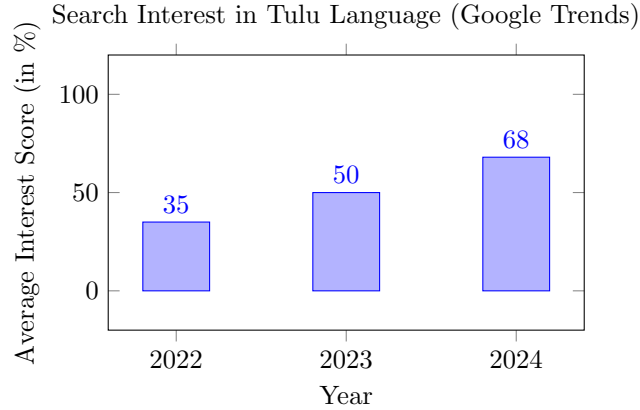
\begin{figure}[htbp]
\centering
\begin{tikzpicture}
\begin{axis}[
    ybar,
    bar width=25pt,
    enlargelimits=0.2,
    ylabel={Average Interest Score (in \%)},
    xlabel={Year},
    symbolic x coords={2022,2023,2024},
    xtick=data,
    ymin=0, ymax=100,
    nodes near coords,
    nodes near coords align={vertical},
    width=0.6\textwidth,
    height=0.4\textwidth,
    title={Search Interest in Tulu Language (Google Trends)}
]
\addplot coordinates {(2022,35) (2023,50) (2024,68)};
\end{axis}
\end{tikzpicture}
\caption{Bar graph showing the rising interest in the Tulu language over the past three years, based on Google Trends \href{https://trends.google.com/trends/explore?q=Tulu}{data}.}
\label{fig:tulu_trend}
\end{figure}

On average, each comment is approximately one sentence long, containing about 8.99 words per sentence. This reflects the informal and concise style typical of social media text. For model development and evaluation, the dataset was divided into training, development, and test sets using the \href{https://tinyurl.com/57h7y87z}{train\_test\_split} 
 function from the scikit-learn library, ensuring a balanced and representative split to support effective model training and evaluation. Each comment was manually categorized into one of the following four mutually exclusive categories:
\begin{itemize}
    \item Not Offensive: Neutral comments without any offensive content.

\item Not Tulu: Comments written in languages other than Tulu, and therefore outside the scope of the target language.

\item Offensive Untargeted: Comments containing offensive language that is not directed at any specific individual or group.

\item Offensive Targeted: Comments that explicitly target an individual or group with offensive or derogatory language.
\end{itemize}

The dataset, illustrated in Figure~\ref{fig:tulu_pie_chart}, is divided into training, development, and test sets, with a distribution of 70\% for training, 15\% for development, and 15\% for testing. The observed class imbalance, especially between offensive and non-offensive categories, mirrors real-world distributions of such content \citep{ataei2022pars}. This emphasises the need for employing effective modelling strategies to tackle this challenge.

\begin{table}[!ht]
\centering
\caption{Linguistic Statistics of the Tulu Offensive Language Dataset}
\label{tab:label_distribution}
\begin{tabular}{lc}
\hline
Metric & Value \\
\hline
Number of Words & 41,384 \\
Vocabulary Size & 13,210 \\
Number of Comments & 3,845 \\
Number of Sentences & 4,604 \\
Average Words per Sentence & 9 \\
Average Sentences per Comment & 1 \\
\hline
\end{tabular}
\end{table}

\section{Methodology}
\subsection{Pre-processing}
The preprocessing phase cleans and standardizes \citep{chakravarthi2020corpus} Tulu text data for machine learning by removing mentions, hashtags, punctuation, digits, extra whitespace, and common English stop words. These steps reduce noise and dimensionality, producing cleaner input for effective training and evaluation of NLP models.
\\
For Example: "Nija ...onjonji padala manas g naatundu .. aven panthinarla aath porludu panther ...song baredinareg music korthinareg bokka aven  panthinareg \adjustbox{valign=c}{\includegraphics[height=2em, width=4em]{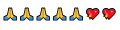}}"
 \\ \textit{Translator: Really... there’s a feeling deep in the heart. Who sang that song so beautifully? And who wrote the song for them}\adjustbox{valign=c}{\includegraphics[height=2em, width=4em]{emorji.drawio.png}}"
\\
 This sentence will be Pre-processed as: "nija onjonji padala manas g naatundu aven panthinarla aath porludu panther song baredinareg music korthinareg bokka aven panthinareg"
\begin{figure}[h]
\centering
\includegraphics[width=1.02\columnwidth]{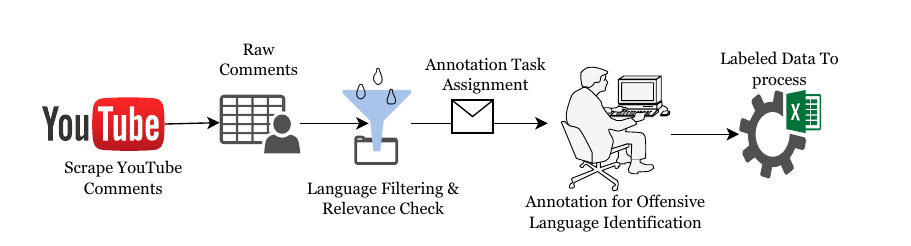} 
\caption{Pipeline depicting the steps involved in dataset creation for OLI from YouTube comments}
\label{figure3}
\end{figure}

\subsection{Feature Extraction}
\begin{itemize}
    \item \textbf{TF-IDF}  is utilized to convert the preprocessed Tulu text into a numerical format suitable for downstream machine learning tasks \citep{rajaraman2011mining}. To capture both local and broader contextual patterns, the vectorizer is configured to include unigrams, bigrams, and trigrams. It is first fitted on the training data, during which it learns the term vocabulary and computes inverse document frequency scores. The same vectorizer is subsequently applied to the development and test sets to ensure consistent feature extraction across all splits. This transformation yields sparse vector representations that quantify the relative importance of terms within individual documents and across the corpus, providing a structured input space for classification models to learn from textual data effectively.
\item \textbf{BERT tokenizer} is used to prepare the text data for input into the BERT model\citep{koroteev2021bert}, tokenization was performed using a pre-trained BERT tokenizer. Each comment was tokenized with truncation and padding applied to maintain a uniform maximum sequence length. The function returns the input\_ids and corresponding attention\_mask, both required by transformer-based models.
\end{itemize}
The deep learning models GRU, BiGRU, LSTM, BiLSTM, and CNN were selected based on their proven effectiveness in handling sequential and code-mixed text in low-resource NLP tasks \citep{zhang2018deep,hande2020kancmd}. Bidirectional and attention-enhanced variants help capture contextual dependencies and improve classification of minority classes. Transformer models such as mBERT and XLM-RoBERTa were included to benchmark performance against state-of-the-art multilingual systems, despite their known limitations on underrepresented languages like Tulu  \citep{chakravarthi2023offensive}.
\subsection{Deep Learning}
This study employs deep learning models with a consistent architectural setup and training configuration. Each model starts with an embedding layer of 100 dimensions, which converts tokenized input text into dense vector representations. After tokenization, the sequence input is padded to a uniform length using Keras' Tokenizer. All recurrent and convolutional layers—including GRU, BiGRU, LSTM, BiLSTM, BiRNN, and Conv1D feature a hidden layer with 128 units. The choice of 128 units balances computational efficiency with sufficient model capacity to capture complex linguistic patterns in code-mixed data, as supported by prior studies that have found this dimensionality effective for text classification tasks in similar domains \citep{Zhang2018, Lai2015, Wang2021}. The final classification layer in each model consists of a dense layer with softmax activation, aimed at predicting one of four categories of offensive language. The Adam optimizer, with a learning rate of 0.001, is utilised during training, and sparse categorical cross-entropy serves as the loss function. Additionally, early stopping and dropout regularization are implemented to prevent overfitting and enhance generalization across all models. Table \ref{tab:dl_performance_grouped} presents the performance analysis of DL models.
\subsubsection{SimpleRNN}
To establish a baseline, a Simple RNN was implemented \citep{sonwani2024simplernn}. This model processes input sequences using a recurrent layer that retains short-term contextual information. While it is computationally efficient, its ability to capture long-range dependencies is limited. The architecture consisted of a SimpleRNN layer with 128 units following the embedding layer, and a dense output layer for classification. Despite its simplicity, this model provided a useful reference point for evaluating more advanced recurrent architectures.

\subsubsection{GRU}
The GRU model was implemented using the Keras Sequential API\citep{zucchet2024recurrent}. It consists of an embedding layer followed by a GRU layer with 100 units to capture temporal dependencies in the input sequences. A dense output layer with softmax activation produces the final class probabilities. The model was compiled using sparse categorical cross-entropy loss and evaluated using accuracy as the performance metric.
 
\subsubsection{Conv1D} A CNN-based architecture was implemented for multi-class text classification\citep{soni2023textconvonet}. Text inputs were tokenized using Keras' Tokenizer and converted to padded sequences with a maximum length derived from the training data. An embedding layer of dimension 100 was used to convert tokens into dense vectors.

The model architecture comprised a Conv1D layer with 128 filters and a kernel size of 5, followed by MaxPooling1D and GlobalMaxPooling1D layers to extract salient features. This was followed by a dense layer with 128 units and ReLU activation, along with a dropout layer (rate =0.5) to reduce overfitting. The final output layer used softmax activation corresponding to the number of classes.
The model was compiled with the Adam optimizer and sparse\_categorical\_crossentropy loss. Early stopping was applied with a patience of 3 to prevent overfitting based on validation loss.

\subsubsection{LSTM}
The architecture includes an embedding layer with a dimension of 100, followed by a LSTM layer with 128 units. A dense output layer with softmax activation was used to predict the class labels. The model was compiled using the Adam optimizer and sparse categorical cross-entropy loss.

\subsubsection{BiLSTM} 
The BiLSTM \citep{liu2019bidirectional} model utilizes a bidirectional Long Short-Term Memory layer to capture contextual dependencies from both the tokens that come before and after each token in the input sequence. This bidirectional design enhances the model's ability to understand the subtle patterns in code-mixed Tulu data, especially when dealing with informal or syntactically variable text.
\subsubsection{BiRNN}
To capture contextual dependencies from both directions in the input text, a BiRNN model was employed using SimpleRNN units \citep{yao2024comparison}. The model architecture consists of an embedding layer with a 100-dimensional vector space, followed by a bidirectional RNN layer with 128 units, and a dense softmax output layer for multi-class classification. The model was compiled using the Adam optimizer with sparse categorical cross-entropy as the loss function.

\subsubsection{BiGRU}

The BiGRU \citep{she2021bigru} model employs a bidirectional Gated Recurrent Unit layer to capture contextual information from both directions of the input sequence. This structure improves the model's capacity to recognize offensive patterns across varying positions in the text, enhancing its effectiveness in sequential modeling tasks involving Tulu code-mixed language.
\subsubsection{BiLSTM + SA}
The model enhances the standard BiLSTM architecture by incorporating a self-attention mechanism \citep{li2020bilstm}. After extracting bidirectional contextual features using the BiLSTM layer, a self-attention layer with an attention size of 256 is applied. This mechanism enables the model to dynamically assign importance to different tokens in the sequence, improving its focus on contextually significant patterns. This combination helps the model better capture subtle and dispersed cues in code-mixed offensive language.
\subsubsection{BiGRU
+SA}
This model extends the BiGRU architecture by integrating a self-attention mechanism that enables the network to focus on the most relevant parts of the input sequence\citep{alayba2018deep}. The self-attention layer computes attention weights over the BiGRU outputs, allowing the model to emphasize informative tokens and suppress irrelevant ones. This is particularly effective for handling noisy, informal, and code-mixed text, resulting in improved classification performance, especially for minority offensive classes.

\begin{table}[!ht]
\centering
\caption{Performance metrics (Precision, Recall, F1-score) for DL models. BiLSTM+SA and BiGRU+SA refer to models enhanced with Self Attention.}
\setlength{\tabcolsep}{6pt}
\renewcommand{\arraystretch}{1.1}

\begin{tabular}{l|ccc|ccc|ccc}
\hline
\multirow{2}{*}{Classes} 
& \multicolumn{3}{c|}{LSTM} 
& \multicolumn{3}{c|}{GRU}
& \multicolumn{3}{c}{SimpleRNN} \\
 & Prec & Rec & F1 & Prec & Rec & F1 & Prec & Rec & F1 \\
\hline
Not Offensive        & 0.47 & 1.00 & 0.64 & 0.47 & 1.00 & 0.64 & 0.81 & 0.71 & 0.76 \\
Not Tulu             & 0.00 & 0.00 & 0.00 & 0.00 & 0.00 & 0.00 & 0.68 & 0.77 & 0.72 \\
Offensive Targeted   & 0.00 & 0.00 & 0.00 & 0.00 & 0.00 & 0.00 & 0.82 & 0.60 & 0.69 \\
Offensive Untargeted & 0.00 & 0.00 & 0.00 & 0.00 & 0.00 & 0.00 & 0.69 & 0.87 & 0.77 \\
\hline

Macro Average        & 0.12 & 0.25  & 0.16 & 0.12 & 0.25  & 0.16 & 0.75 & 0.74 & 0.73 \\
Weighted Average     & 0.22 & 0.47  & 0.30 & 0.22 & 0.47  & 0.30 & 0.75 & 0.74 & 0.74 \\
\hline
\end{tabular}

\vspace{0.5cm}

\begin{tabular}{l|ccc|ccc|ccc}
\hline
\multirow{2}{*}{Classes} 
& \multicolumn{3}{c|}{BiLSTM} 
& \multicolumn{3}{c|}{1D Conv}
& \multicolumn{3}{c}{BiGRU} \\
 & Prec & Rec & F1 & Prec & Rec & F1 & Prec & Rec & F1 \\
\hline
Not Offensive        & 0.84 & 0.81 & 0.83 & 0.84 & 0.81 & 0.83 & 0.86 & 0.76 & 0.81 \\
Not Tulu             & 0.81 & 0.75 & 0.78 & 0.81 & 0.75 & 0.78 & 0.81 & 0.81 & 0.81 \\
Offensive Targeted   & 0.64 & 0.62 & 0.63 & 0.64 & 0.62 & 0.63 & 0.82 & 0.60 & 0.69 \\
Offensive Untargeted & 0.77 & 0.95 & 0.85 & 0.77 & 0.95 & 0.85 & 0.69 & 0.87 & 0.77 \\
\hline

Macro Average        & 0.77 & 0.78  & 0.77 & 0.77 & 0.78  & 0.77 & 0.75 & 0.74 & 0.73 \\
Weighted Average     & 0.80 & 0.80  & 0.80 & 0.80 & 0.80  & 0.80 & 0.75 & 0.74 & 0.74 \\
\hline
\end{tabular}

\vspace{0.5cm}

\begin{tabular}{l|ccc|ccc|ccc}
\hline
\multirow{2}{*}{Classes} 
& \multicolumn{3}{c|}{BiRNN} 
& \multicolumn{3}{c|}{BiLSTM+ A}
& \multicolumn{3}{c}{BiGRU+SA} \\
 & Prec & Rec & F1 & Prec & Rec & F1 & Prec & Rec & F1 \\
\hline
Not Offensive        & 0.81 & 0.71 & 0.76 & 0.85 & 0.88 & 0.87 & 0.81 & 0.85 & 0.83 \\
Not Tulu             & 0.78 & 0.78 & 0.78 & 0.73 & 0.78 & 0.78 & 0.73 & 0.78 & 0.78 \\
Offensive Targeted   & 0.73 & 0.62 & 0.67 & 0.75 & 0.06 & 0.11 & 0.74 & 0.77 & 0.75 \\
Offensive Untargeted & 0.74 & 0.92 & 0.82 & 0.00 & 0.00 & 0.00 & 0.85 & 0.88 & 0.87 \\
\hline
Macro Average        & 0.78 & 0.78  & 0.77 & 0.32 & 0.37  & 0.33 & 0.81 & 0.81 & 0.81 \\
Weighted Average     & 0.80 & 0.80  & 0.80 & 0.45 & 0.58  & 0.49 & 0.82 & 0.82 & 0.81 \\
\hline
\end{tabular}

\label{tab:dl_performance_grouped}
\end{table}

\subsection{Transfer Learning Models}
Transfer learning with pretrained multilingual transformer models has shown strong performance in various natural language processing tasks by utilizing extensive corpora. However, these models often perform poorly on low-resource languages like Tulu, mainly due to inadequate representation of such languages in their pretraining data. We conducted experiments with multilingual pretrained models, including mBERT and XLM-RoBERTa, to assess their effectiveness in identifying offensive language in code-mixed Tulu text. The results underscore the challenges that transfer learning approaches face when applied to underrepresented languages. Table~\ref{tab:tl_performance} presents the performance analysis of mBERT and XLM-RoBERTa.
\subsubsection{mBert}
The model accepts tokenized text (input\_ids) and corresponding attention masks as input, which are processed by the mBERT layer to generate contextualized word embeddings\citep{ragab2025multilingual}. These embeddings are aggregated using a Global Average Pooling layer to obtain a fixed-size representation, followed by a Dropout layer to prevent overfitting. The final classification is performed using a Dense layer with softmax activation to output class probabilities. The model is compiled using the Adam optimizer and sparse categorical cross-entropy loss, making it effective for classifying text across diverse languages. 

\subsubsection{XLM-RoBERTa}
The XLM-RoBERTa-based classifier utilizes tokenized input and attention masks to extract multilingual contextual features via a custom XLM-RoBERTa layer. These features are aggregated using Global Average Pooling, followed by dropout for regularization and a dense softmax layer for final classification. This architecture is well-suited for robust multilingual text classification tasks.
\section{Experiments and Results }
\subsection{Experimental Setup}

The experiments were conducted using \href{https://shorturl.at/qbE1Y}{Google Colab}
 with a Tesla T4 GPU environment. The dataset comprised Tulu-Kannada-English  code-mixed YouTube comments, annotated into four categories: Not Offensive, Not Tulu, Offensive Targeted, and Offensive Untargeted. Preprocessing steps included tokenization, lowercasing, and padding sequences to a uniform length. 

While pre-trained embeddings like FastText \citep{bojanowski2017enriching} and GloVe \cite{pennington2014glove} are widely used for resource-rich languages, such embeddings are currently unavailable for Tulu due to its under-resourced nature. As a result, the model relied on randomly initialized word vectors to represent the vocabulary.

The models were implemented using TensorFlow and Keras frameworks. Training was performed using the Adam optimizer with a learning rate of 0.001 and categorical cross-entropy as the loss function. Early stopping and dropout regularization were employed to prevent overfitting.

\subsection{Results analysis}
The experimental results for the code-mixed Tulu dataset were analyzed across multiple deep learning architectures. These include traditional RNN-based models (LSTM, GRU, SimpleRNN), their bidirectional variants (BiLSTM, BiGRU, BiRNN), CNN-based architectures, and attention-augmented variants, along with transformer-based baselines such as mBERT and XLM-Roberta. Performance was evaluated using precision, recall, F1-score, and accuracy across four target classes: Not Offensive, Not Tulu, Offensive Targeted, and Offensive Untargeted.

Among the tested models, BiGRU +SA outperformed others, achieving an accuracy of 82\%, with a macro F1-score of 0.81, highlighting its robustness in learning contextual dependencies and offensive nuance across under-resourced language data. In contrast, basic LSTM and GRU models showed significantly lower performance (accuracy 47\%), mainly due to their inability to capture complex semantic relationships, particularly in the minority classes. As shown in Table 2 and Table 3, the performance comparison of deep learning and transformer models highlights their effectiveness in Offensive Language Identification for Tulu.

\subsection{Class-Wise Evaluation and Minority Class Handling
}
The dataset exhibited a class imbalance, where the "Not Offensive" class dominated the distribution. Simple models like LSTM and GRU performed adequately only for this majority class (recall = 1.00), while they entirely failed to predict instances from the "Offensive Targeted" and "Offensive Untargeted" classes (precision, recall = 0.00). These results emphasize that standard recurrent models are highly biased toward frequent classes under low-resource settings.

In contrast, attention-based architectures (BiGRU + SA, BiLSTM + SA) demonstrated better generalization across all classes, with F1-scores above 0.75 even for minority categories. \cite{hande2020kancmd} supports findings in the literature  that incorporating attention enhances focus on informative tokens, especially crucial in code-mixed and offensive content \citep{bahdanau2015neural}.

\begin{table}[ht]
\centering
\caption{Performance metrics (Precision, Recall, F1-score) for Transfer Learning models (mBERT and XLM-Roberta).}
\setlength{\tabcolsep}{9pt}
\renewcommand{\arraystretch}{1.1}

\begin{tabular}{l|ccc|ccc}
\hline
\multirow{2}{*}{Classes} 
& \multicolumn{3}{c|}{mBERT} 
& \multicolumn{3}{c}{XLMRoberta} \\
 & Prec & Rec & F1 & Prec & Rec & F1 \\
\hline
Not Offensive        & 0.80 & 0.83 & 0.68 & 0.83 & 0.75 & 0.54 \\
Not Tulu             & 0.75 & 0.75 & 0.75 & 0.73 & 0.56 & 0.63 \\
Offensive Targeted   & 0.75 & 0.78 & 0.73 & 0.00 & 0.00 & 0.00 \\
Offensive Untargeted & 0.53 & 0.46 & 0.49 & 0.00 & 0.00 & 0.00 \\
\hline
Macro Average        & 0.68 & 0.53  & 0.53 & 0.32 & 0.37  & 0.33 \\
Weighted Average     & 0.68 & 0.68  & 0.65 & 0.45 & 0.58  & 0.49 \\
\hline
\end{tabular}
\label{tab:tl_performance}
\end{table}

\subsubsection{Comparative Evaluation with Transformer-Based Models}
Transformer baselines such as mBERT and XLM-Roberta provided moderate performance with accuracies of 68\% and 58\%, respectively. While mBERT achieved high precision on the "Not Tulu" class (0.75), it struggled with "Offensive Targeted" (F1 = 0.11), indicating insufficient contextual disambiguation for highly nuanced Targeted offensive content. XLM-Roberta further underperformed, especially on offensive categories, returning zero F1-scores for both Offensive Targeted and     
Offensive Untargeted 
 classes.

\par These observations align with recent findings that multilingual transformers, though trained on extensive corpora, are less effective on code-mixed, under-resourced language pairs without task-specific fine-tuning or domain adaptation.

\subsection{Impact of Model Architecture and Attention Mechanisms}
The use of bidirectional context and self attention mechanisms significantly improved model effectiveness. BiGRU + SA achieved an F1-score of 0.87 on the "Offensive Untargeted" class, a class often misclassified in simpler models. The model’s capability to attend to both backward and forward contexts with self-attention layers helped isolate offensive patterns irrespective of position or spelling variation.
Conv1D also achieved competitive results (80\% accuracy), comparable to BiLSTM, suggesting that local n-gram level features remain relevant for certain offensive patterns.

\section{Discussion}
This study explored various deep learning architectures and transformer-based models to detect offensive language in Tulu, a low-resource and code-mixed language. The findings highlight the strengths and limitations of these models in handling the unique challenges posed by Tulu's linguistic characteristics.

\subsection{Model Performance Comparison}

The results highlight significant variations in the effectiveness of different deep learning models for offensive language detection in Tulu. Among traditional architectures, the BiGRU with self-attention emerged as the most robust model, achieving a balanced performance across multiple categories. It excelled in identifying the “Offensive Untargeted” class, attaining a high recall of 0.92 and an F1-score of 0.82. This suggests that the model is adept at detecting general offensive language that is not specifically targeted.

Other strong performers included BiLSTM, 1D CNN, and BiGRU without self-attention, which demonstrated solid results particularly in distinguishing non-offensive classes such as “Not Offensive” and “Not Tulu.” These models showed strength in classifying broader categories but were less effective in capturing subtler, targeted offensive content. Their performance on the “Offensive Targeted” class was suboptimal, indicating the need for further model improvements to handle nuanced offensive expressions.

Interestingly, the SimpleRNN model showed comparatively better results than some more complex architectures, suggesting that in low-resource scenarios, simpler models with effective feature extraction can sometimes outperform deeper networks. However, the combination of bidirectional sequence modeling and self-attention in the BiGRU architecture distinctly outperformed all others, confirming the advantage of leveraging both contextual directions alongside attention mechanisms to capture intricate patterns in the text.
\subsection{Transfer Learning vs Bidirectional Setup}
A pronounced performance gap was observed between transformer-based transfer learning models and traditional deep learning models on the Tulu dataset. Both mBERT \citep{devlin2019bert} and XLM-RoBERTa \cite{conneau2020unsupervised}, pretrained on extensive multilingual corpora spanning dozens of languages, performed moderately at best, with accuracies of 0.68 and 0.58 respectively. Despite their general-purpose multilingual capabilities, these models struggled with Tulu, a language notably underrepresented in their pretraining datasets.

Specifically, mBERT’s F1-score for the “Offensive Targeted” class was as low as 0.11, revealing limited effectiveness in capturing Tulu-specific syntactic and lexical features. XLM-RoBERTa’s performance was even more constrained, achieving near-zero F1-scores for both offensive classes. These results underscore the inherent challenges transfer learning models face when applied to low-resource, code-mixed languages with distinctive morphology and syntax, such as Tulu.

The difficulty arises primarily due to the domain gap: transfer learning models thrive when large-scale, high-quality annotated data are available, but struggle with languages lacking such resources. Fine-tuning on domain-specific datasets can mitigate this issue; however, the scarcity of labeled Tulu data poses a significant barrier. Moreover, Tulu’s unique linguistic characteristics and code-switching tendencies further complicate generalization by pretrained models.

Consequently, this study emphasizes the need for developing specialized datasets and customized architectures for under-resourced languages. While transfer learning remains a promising approach, its success is contingent on the availability of representative data and effective fine-tuning strategies tailored to the target language and domain.

\section{Conclusion}
This paper presents the creation of a novel Tulu corpus for OLI, contributing to the field of low-resource languages and multilingual NLP. The corpus, annotated with four categories: Not Offensive, Not Tulu, Offensive Targeted, and Offensive Untargeted addresses a significant gap in resources for Dravidian languages. Several deep learning models, including GRU, BiGRU, BiLSTM, and CNN, were evaluated, with BiGRU integrated with a self-attention mechanism outperforming others, achieving an accuracy of 82\%. This result highlights the effectiveness of bidirectional sequence modeling combined with attention mechanisms in capturing complex contextual and syntactic patterns in low-resource languages like Tulu. \par The study also demonstrated the challenges of using pre-trained transformer models, such as mBERT and XLM-Roberta, for Tulu due to their lack of representation in pre-training corpora. These findings emphasize the importance of designing language-specific models and corpora, as well as the need to fine-tune pre-trained models for under-resourced languages.
Future research should focus on expanding corpora with diverse text sources and dialects to enhance model generalization. Fine-tuning transformer-based models like mBERT, XLM-Roberta, or exploring newer architectures such as T5 and GPT-3 could improve performance. Incorporating cross-lingual transfer learning and additional linguistic features like morphology and sentiment may further boost accuracy. Developing real-time offensive language detection tools can also support practical NLP applications. Advancing resources for low-resource languages remains essential to ensure inclusive and impactful NLP technologies.

\section*{Acknowledgements}
The authors would like to sincerely thank Dr. Dinakara P and Mr. Niyaz Padil from the Department of Linguistics, Yenepoya (Deemed to be University) for their valuable contributions to the annotation of the TuluOffense dataset. Their linguistic expertise and commitment were instrumental in the creation of this resource.


\begin{thebibliography}{40}
\ifx \bisbn   \undefined \def \bisbn  #1{ISBN #1}\fi
\ifx \binits  \undefined \def \binits#1{#1}\fi
\ifx \bauthor  \undefined \def \bauthor#1{#1}\fi
\ifx \batitle  \undefined \def \batitle#1{#1}\fi
\ifx \bjtitle  \undefined \def \bjtitle#1{#1}\fi
\ifx \bvolume  \undefined \def \bvolume#1{\textbf{#1}}\fi
\ifx \byear  \undefined \def \byear#1{#1}\fi
\ifx \bissue  \undefined \def \bissue#1{#1}\fi
\ifx \bfpage  \undefined \def \bfpage#1{#1}\fi
\ifx \blpage  \undefined \def \blpage #1{#1}\fi
\ifx \burl  \undefined \def \burl#1{\textsf{#1}}\fi
\ifx \doiurl  \undefined \def \doiurl#1{\url{https://doi.org/#1}}\fi
\ifx \betal  \undefined \def \betal{\textit{et al.}}\fi
\ifx \binstitute  \undefined \def \binstitute#1{#1}\fi
\ifx \binstitutionaled  \undefined \def \binstitutionaled#1{#1}\fi
\ifx \bctitle  \undefined \def \bctitle#1{#1}\fi
\ifx \beditor  \undefined \def \beditor#1{#1}\fi
\ifx \bpublisher  \undefined \def \bpublisher#1{#1}\fi
\ifx \bbtitle  \undefined \def \bbtitle#1{#1}\fi
\ifx \bedition  \undefined \def \bedition#1{#1}\fi
\ifx \bseriesno  \undefined \def \bseriesno#1{#1}\fi
\ifx \blocation  \undefined \def \blocation#1{#1}\fi
\ifx \bsertitle  \undefined \def \bsertitle#1{#1}\fi
\ifx \bsnm \undefined \def \bsnm#1{#1}\fi
\ifx \bsuffix \undefined \def \bsuffix#1{#1}\fi
\ifx \bparticle \undefined \def \bparticle#1{#1}\fi
\ifx \barticle \undefined \def \barticle#1{#1}\fi
\bibcommenthead
\ifx \bconfdate \undefined \def \bconfdate #1{#1}\fi
\ifx \botherref \undefined \def \botherref #1{#1}\fi
\ifx \url \undefined \def \url#1{\textsf{#1}}\fi
\ifx \bchapter \undefined \def \bchapter#1{#1}\fi
\ifx \bbook \undefined \def \bbook#1{#1}\fi
\ifx \bcomment \undefined \def \bcomment#1{#1}\fi
\ifx \oauthor \undefined \def \oauthor#1{#1}\fi
\ifx \citeauthoryear \undefined \def \citeauthoryear#1{#1}\fi
\ifx \endbibitem  \undefined \def \endbibitem {}\fi
\ifx \bconflocation  \undefined \def \bconflocation#1{#1}\fi
\ifx \arxivurl  \undefined \def \arxivurl#1{\textsf{#1}}\fi
\csname PreBibitemsHook\endcsname

\bibitem[\protect\citeauthoryear{Alayba et~al.}{2018}]{alayba2018deep}
\begin{bchapter}
\bauthor{\bsnm{Alayba}, \binits{S.}},
\bauthor{\bsnm{Li}, \binits{Y.}},
\bauthor{\bsnm{Sutinen}, \binits{E.}},
\bauthor{\bsnm{Rayson}, \binits{P.}}:
\bctitle{Deep learning for sentiment analysis: A comparative study}.
In: \bbtitle{2018 4th International Conference on Information Management (ICIM)},
pp. \bfpage{87}--\blpage{92}
(\byear{2018}).
\doiurl{10.1109/INFOMAN.2018.8392806} .
\bcomment{IEEE}
\end{bchapter}
\endbibitem

\bibitem[\protect\citeauthoryear{Bahdanau et~al.}{2015}]{bahdanau2015neural}
\begin{bchapter}
\bauthor{\bsnm{Bahdanau}, \binits{D.}},
\bauthor{\bsnm{Cho}, \binits{K.}},
\bauthor{\bsnm{Bengio}, \binits{Y.}}:
\bctitle{Neural machine translation by jointly learning to align and translate}.
In: \bbtitle{3rd International Conference on Learning Representations (ICLR)}
(\byear{2015}).
\burl{https://arxiv.org/abs/1409.0473}
\end{bchapter}
\endbibitem

\bibitem[\protect\citeauthoryear{Bojanowski et~al.}{2017}]{bojanowski2017enriching}
\begin{bchapter}
\bauthor{\bsnm{Bojanowski}, \binits{P.}},
\bauthor{\bsnm{Grave}, \binits{E.}},
\bauthor{\bsnm{Joulin}, \binits{A.}},
\bauthor{\bsnm{Mikolov}, \binits{T.}}:
\bctitle{Enriching word vectors with subword information}.
In: \bbtitle{Proceedings of the 2017 Conference on Empirical Methods in Natural Language Processing (EMNLP)},
pp. \bfpage{1}--\blpage{10}
(\byear{2017})
\end{bchapter}
\endbibitem

\bibitem[\protect\citeauthoryear{Bhat}{2019}]{bhat2019tulu}
\begin{bchapter}
\bauthor{\bsnm{Bhat}, \binits{D.S.}}:
\bctitle{Tulu}.
In: \bbtitle{The Dravidian Languages},
pp. \bfpage{219}--\blpage{236}.
\bpublisher{Routledge}, \blocation{???}
(\byear{2019})
\end{bchapter}
\endbibitem

\bibitem[\protect\citeauthoryear{Chakravarthi et~al.}{2023}]{chakravarthi2023offensive}
\begin{barticle}
\bauthor{\bsnm{Chakravarthi}, \binits{B.R.}},
\bauthor{\bsnm{Jagadeeshan}, \binits{M.B.}},
\bauthor{\bsnm{Palanikumar}, \binits{V.}},
\bauthor{\bsnm{Priyadharshini}, \binits{R.}}:
\batitle{Offensive language identification in dravidian languages using mpnet and cnn}.
\bjtitle{International Journal of Information Management Data Insights}
\bvolume{3}(\bissue{1}),
\bfpage{100151}
(\byear{2023})
\end{barticle}
\endbibitem

\bibitem[\protect\citeauthoryear{Conneau et~al.}{2020}]{conneau2020unsupervised}
\begin{botherref}
\oauthor{\bsnm{Conneau}, \binits{A.}},
\oauthor{\bsnm{Khandelwal}, \binits{K.}},
\oauthor{\bsnm{Goyal}, \binits{N.}},
\oauthor{\bsnm{Chaudhary}, \binits{V.}},
\oauthor{\bsnm{Wenzek}, \binits{G.}},
\oauthor{\bsnm{Guzmán}, \binits{F.}},
\oauthor{\bsnm{Grave}, \binits{E.}},
\oauthor{\bsnm{Ott}, \binits{M.}},
\oauthor{\bsnm{Zettlemoyer}, \binits{L.}},
\oauthor{\bsnm{Stoyanov}, \binits{V.}}:
Unsupervised cross-lingual representation learning at scale.
arXiv preprint arXiv:1911.02116
(2020)
\end{botherref}
\endbibitem

\bibitem[\protect\citeauthoryear{Chakravarthi et~al.}{2020}]{chakravarthi2020corpus}
\begin{botherref}
\oauthor{\bsnm{Chakravarthi}, \binits{B.R.}},
\oauthor{\bsnm{Muralidaran}, \binits{V.}},
\oauthor{\bsnm{Priyadharshini}, \binits{R.}},
\oauthor{\bsnm{McCrae}, \binits{J.P.}}:
Corpus creation for sentiment analysis in code-mixed tamil-english text.
arXiv preprint arXiv:2006.00206
(2020)
\end{botherref}
\endbibitem

\bibitem[\protect\citeauthoryear{Chakravarthi et~al.}{2022}]{chakravarthi2022dravidiancodemix}
\begin{barticle}
\bauthor{\bsnm{Chakravarthi}, \binits{B.R.}},
\bauthor{\bsnm{Priyadharshini}, \binits{R.}},
\bauthor{\bsnm{Muralidaran}, \binits{V.}},
\bauthor{\bsnm{Jose}, \binits{N.}},
\bauthor{\bsnm{Suryawanshi}, \binits{S.}},
\bauthor{\bsnm{Sherly}, \binits{E.}},
\bauthor{\bsnm{McCrae}, \binits{J.P.}}:
\batitle{Dravidiancodemix: Sentiment analysis and offensive language identification dataset for dravidian languages in code-mixed text}.
\bjtitle{Language Resources and Evaluation}
\bvolume{56}(\bissue{3}),
\bfpage{765}--\blpage{806}
(\byear{2022})
\end{barticle}
\endbibitem

\bibitem[\protect\citeauthoryear{Durairaj et~al.}{2025}]{durairaj2025}
\begin{bchapter}
\bauthor{\bsnm{Durairaj}, \binits{T.}},
\bauthor{\bsnm{Chakravarthi}, \binits{B.R.}},
\bauthor{\bsnm{Hegde}, \binits{A.}},
\bauthor{\bsnm{Shashirekha}, \binits{H.L.}},
\bauthor{\bsnm{Natarajan}, \binits{R.}},
\bauthor{\bsnm{Thavareesan}, \binits{S.}},
\bauthor{\bsnm{Sakuntharaj}, \binits{R.}},
\bauthor{\bsnm{Rajkumar}, \binits{C.}},
\bauthor{\bsnm{Shetty}, \binits{P.}},
\bauthor{\bsnm{Kumar}, \binits{H.S.}}, \betal:
\bctitle{Overview of the shared task on sentiment analysis in tamil and tulu}.
In: \bbtitle{Proceedings of the Fifth Workshop on Speech, Vision, and Language Technologies for Dravidian Languages},
pp. \bfpage{732}--\blpage{738}
(\byear{2025})
\end{bchapter}
\endbibitem

\bibitem[\protect\citeauthoryear{Devlin et~al.}{2019}]{devlin2019bert}
\begin{botherref}
\oauthor{\bsnm{Devlin}, \binits{J.}},
\oauthor{\bsnm{Chang}, \binits{M.-W.}},
\oauthor{\bsnm{Lee}, \binits{K.}},
\oauthor{\bsnm{Toutanova}, \binits{K.}}:
Bert: Pre-training of deep bidirectional transformers for language understanding.
arXiv preprint arXiv:1810.04805
(2019)
\end{botherref}
\endbibitem

\bibitem[\protect\citeauthoryear{Garg and Sharma}{2020}]{garg2020annotated}
\begin{barticle}
\bauthor{\bsnm{Garg}, \binits{N.}},
\bauthor{\bsnm{Sharma}, \binits{K.}}:
\batitle{Annotated corpus creation for sentiment analysis in code-mixed hindi-english (hinglish) social network data}.
\bjtitle{Indian Journal of Science and Technology}
\bvolume{13}(\bissue{40}),
\bfpage{4216}--\blpage{4224}
(\byear{2020})
\end{barticle}
\endbibitem

\bibitem[\protect\citeauthoryear{Hegde et~al.}{2022a}]{hegde2022}
\begin{bchapter}
\bauthor{\bsnm{Hegde}, \binits{A.}},
\bauthor{\bsnm{Anusha}, \binits{M.D.}},
\bauthor{\bsnm{Coelho}, \binits{S.}},
\bauthor{\bsnm{Shashirekha}, \binits{H.L.}},
\bauthor{\bsnm{Chakravarthi}, \binits{B.R.}}:
\bctitle{Corpus creation for sentiment analysis in code-mixed tulu text}.
In: \bbtitle{Proceedings of the 1st Annual Meeting of the ELRA/ISCA Special Interest Group on Under-Resourced Languages},
pp. \bfpage{33}--\blpage{40}
(\byear{2022})
\end{bchapter}
\endbibitem

\bibitem[\protect\citeauthoryear{Hegde et~al.}{2022b}]{hegde2022corpus}
\begin{bchapter}
\bauthor{\bsnm{Hegde}, \binits{A.}},
\bauthor{\bsnm{Anusha}, \binits{M.D.}},
\bauthor{\bsnm{Coelho}, \binits{S.}},
\bauthor{\bsnm{Shashirekha}, \binits{H.L.}},
\bauthor{\bsnm{Chakravarthi}, \binits{B.R.}}:
\bctitle{Corpus creation for sentiment analysis in code-mixed tulu text}.
In: \bbtitle{Proceedings of the 1st Annual Meeting of the ELRA/ISCA Special Interest Group on Under-resourced Languages},
pp. \bfpage{33}--\blpage{40}
(\byear{2022})
\end{bchapter}
\endbibitem

\bibitem[\protect\citeauthoryear{Hegde and Lakshmaiah}{2023}]{asha2023kt2}
\begin{bchapter}
\bauthor{\bsnm{Hegde}, \binits{A.}},
\bauthor{\bsnm{Lakshmaiah}, \binits{S.H.}}:
\bctitle{Kt2: Kannada-tulu parallel corpus construction for neural machine translation}.
In: \bbtitle{Proceedings of the 20th International Conference on Natural Language Processing (ICON)},
pp. \bfpage{743}--\blpage{753}
(\byear{2023})
\end{bchapter}
\endbibitem

\bibitem[\protect\citeauthoryear{Hande et~al.}{2020}]{hande2020kancmd}
\begin{bchapter}
\bauthor{\bsnm{Hande}, \binits{A.}},
\bauthor{\bsnm{Priyadharshini}, \binits{R.}},
\bauthor{\bsnm{Chakravarthi}, \binits{B.R.}}:
\bctitle{Kancmd: Kannada codemixed dataset for sentiment analysis and offensive language detection}.
In: \bbtitle{Proceedings of the Third Workshop on Computational Modeling of People’s Opinions, Personality, and Emotion’s in Social Media},
pp. \bfpage{54}--\blpage{63}
(\byear{2020})
\end{bchapter}
\endbibitem

\bibitem[\protect\citeauthoryear{Hu et~al.}{2020}]{hu2020xtreme}
\begin{bchapter}
\bauthor{\bsnm{Hu}, \binits{J.}},
\bauthor{\bsnm{Ruder}, \binits{S.}},
\bauthor{\bsnm{Siddhant}, \binits{A.}},
\bauthor{\bsnm{Neubig}, \binits{G.}},
\bauthor{\bsnm{Firat}, \binits{O.}},
\bauthor{\bsnm{Johnson}, \binits{M.}}:
\bctitle{Xtreme: A massively multilingual multi-task benchmark for evaluating cross-lingual generalization}.
In: \bbtitle{Proceedings of the 37th International Conference on Machine Learning},
pp. \bfpage{4411}--\blpage{4421}
(\byear{2020})
\end{bchapter}
\endbibitem

\bibitem[\protect\citeauthoryear{Joshi et~al.}{2016}]{joshi2016towards}
\begin{bchapter}
\bauthor{\bsnm{Joshi}, \binits{A.}},
\bauthor{\bsnm{Bhattacharyya}, \binits{P.}},
\bauthor{\bsnm{Carman}, \binits{M.J.}}:
\bctitle{Towards sub-word level compositions for sentiment analysis of hindi-english code mixed text}.
In: \bbtitle{Proceedings of the 2016 Conference on Empirical Methods in Natural Language Processing},
pp. \bfpage{158}--\blpage{163}
(\byear{2016})
\end{bchapter}
\endbibitem

\bibitem[\protect\citeauthoryear{Kulkarni-Joshi}{2019}]{kulkarni2019linguistic}
\begin{barticle}
\bauthor{\bsnm{Kulkarni-Joshi}, \binits{S.}}:
\batitle{Linguistic history and language diversity in india: Views and counterviews}.
\bjtitle{Journal of Biosciences}
\bvolume{44}(\bissue{3}),
\bfpage{62}
(\byear{2019})
\end{barticle}
\endbibitem

\bibitem[\protect\citeauthoryear{Koroteev}{2021}]{koroteev2021bert}
\begin{botherref}
\oauthor{\bsnm{Koroteev}, \binits{M.V.}}:
Bert: a review of applications in natural language processing and understanding.
arXiv preprint arXiv:2103.11943
(2021)
\end{botherref}
\endbibitem

\bibitem[\protect\citeauthoryear{Krippendorff}{2011}]{krippendorff2011agreement}
\begin{barticle}
\bauthor{\bsnm{Krippendorff}, \binits{K.}}:
\batitle{Agreement and information in the reliability of coding}.
\bjtitle{Communication Methods and Measures}
\bvolume{5}(\bissue{2}),
\bfpage{93}--\blpage{112}
(\byear{2011})
\end{barticle}
\endbibitem

\bibitem[\protect\citeauthoryear{Liu and Guo}{2019}]{liu2019bidirectional}
\begin{barticle}
\bauthor{\bsnm{Liu}, \binits{G.}},
\bauthor{\bsnm{Guo}, \binits{J.}}:
\batitle{Bidirectional lstm with attention mechanism and convolutional layer for text classification}.
\bjtitle{Neurocomputing}
\bvolume{337},
\bfpage{325}--\blpage{338}
(\byear{2019})
\end{barticle}
\endbibitem

\bibitem[\protect\citeauthoryear{Li et~al.}{2020}]{li2020bilstm}
\begin{barticle}
\bauthor{\bsnm{Li}, \binits{W.}},
\bauthor{\bsnm{Qi}, \binits{F.}},
\bauthor{\bsnm{Yu}, \binits{Z.}}:
\batitle{Bidirectional lstm with self-attention mechanism and multi-channel features for sentiment classification}.
\bjtitle{Neurocomputing}
\bvolume{387},
\bfpage{63}--\blpage{77}
(\byear{2020})
\doiurl{10.1016/j.neucom.2020.01.006}
\end{barticle}
\endbibitem

\bibitem[\protect\citeauthoryear{Lai et~al.}{2015}]{Lai2015}
\begin{bchapter}
\bauthor{\bsnm{Lai}, \binits{S.}},
\bauthor{\bsnm{Xu}, \binits{L.}},
\bauthor{\bsnm{Liu}, \binits{K.}},
\bauthor{\bsnm{Zhao}, \binits{J.}}:
\bctitle{Recurrent convolutional neural networks for text classification}.
In: \bbtitle{Proceedings of the AAAI Conference on Artificial Intelligence},
pp. \bfpage{2267}--\blpage{2273}
(\byear{2015})
\end{bchapter}
\endbibitem

\bibitem[\protect\citeauthoryear{Pennington et~al.}{2014}]{pennington2014glove}
\begin{bchapter}
\bauthor{\bsnm{Pennington}, \binits{J.}},
\bauthor{\bsnm{Socher}, \binits{R.}},
\bauthor{\bsnm{Manning}, \binits{C.D.}}:
\bctitle{Glove: Global vectors for word representation}.
In: \bbtitle{Proceedings of the 2014 Conference on Empirical Methods in Natural Language Processing (EMNLP)},
pp. \bfpage{1532}--\blpage{1543}
(\byear{2014})
\end{bchapter}
\endbibitem

\bibitem[\protect\citeauthoryear{Ragab et~al.}{2025}]{ragab2025multilingual}
\begin{bchapter}
\bauthor{\bsnm{Ragab}, \binits{M.I.}},
\bauthor{\bsnm{Mohamed}, \binits{E.H.}},
\bauthor{\bsnm{Medhat}, \binits{W.}}:
\bctitle{Multilingual propaganda detection: Exploring transformer-based models mbert, xlm-roberta, and mt5}.
In: \bbtitle{Proceedings of the First International Workshop on Nakba Narratives as Language Resources},
pp. \bfpage{75}--\blpage{82}
(\byear{2025})
\end{bchapter}
\endbibitem

\bibitem[\protect\citeauthoryear{Rajaraman and Ullman}{2011}]{rajaraman2011mining}
\begin{bbook}
\bauthor{\bsnm{Rajaraman}, \binits{A.}},
\bauthor{\bsnm{Ullman}, \binits{J.D.}}:
\bbtitle{Mining of Massive Datasets},
\bedition{1st} edn.
\bpublisher{Cambridge University Press}, \blocation{???}
(\byear{2011})
\end{bbook}
\endbibitem

\bibitem[\protect\citeauthoryear{Shangipour~Ataei et~al.}{2022}]{ataei2022pars}
\begin{bchapter}
\bauthor{\bsnm{Shangipour~Ataei}, \binits{T.}},
\bauthor{\bsnm{Darvishi}, \binits{K.}},
\bauthor{\bsnm{Javdan}, \binits{S.}},
\bauthor{\bsnm{Minaei-Bidgoli}, \binits{B.}},
\bauthor{\bsnm{Eetemadi}, \binits{S.}}:
\bctitle{Pars-absa: a manually annotated aspect-based sentiment analysis benchmark on farsi product reviews}.
In: \bbtitle{Proceedings of the Thirteenth Language Resources and Evaluation Conference},
pp. \bfpage{7056}--\blpage{7060}.
\bpublisher{European Language Resources Association},
\blocation{Marseille, France}
(\byear{2022}).
\burl{https://aclanthology.org/2022.lrec-1.763/}
\end{bchapter}
\endbibitem

\bibitem[\protect\citeauthoryear{Sonwani et~al.}{2024}]{sonwani2024simplernn}
\begin{bchapter}
\bauthor{\bsnm{Sonwani}, \binits{H.}},
\bauthor{\bsnm{Banoth}, \binits{E.}},
\bauthor{\bsnm{Jain}, \binits{P.K.}}:
\bctitle{Simplernn based human emotion recognition using eeg signals}.
In: \bbtitle{International Conference on Computational Intelligence in Communications and Business Analytics},
pp. \bfpage{48}--\blpage{57}.
\bpublisher{Springer}, \blocation{???}
(\byear{2024})
\end{bchapter}
\endbibitem

\bibitem[\protect\citeauthoryear{Suryawanshi et~al.}{2020}]{suryawanshi2020multimodal}
\begin{bchapter}
\bauthor{\bsnm{Suryawanshi}, \binits{S.}},
\bauthor{\bsnm{Chakravarthi}, \binits{B.R.}},
\bauthor{\bsnm{Priyadharshini}, \binits{R.}},
\bauthor{\bsnm{Sherly}, \binits{E.}}:
\bctitle{A multimodal approach to detect offensive language in dravidian code-mixed youtube comments}.
In: \bbtitle{Proceedings of the Second Workshop on Trolling, Aggression and Cyberbullying (TRAC-2)},
pp. \bfpage{16}--\blpage{24}
(\byear{2020})
\end{bchapter}
\endbibitem

\bibitem[\protect\citeauthoryear{Soni et~al.}{2023}]{soni2023textconvonet}
\begin{barticle}
\bauthor{\bsnm{Soni}, \binits{S.}},
\bauthor{\bsnm{Chouhan}, \binits{S.S.}},
\bauthor{\bsnm{Rathore}, \binits{S.S.}}:
\batitle{Textconvonet: A convolutional neural network based architecture for text classification}.
\bjtitle{Applied Intelligence}
\bvolume{53}(\bissue{11}),
\bfpage{14249}--\blpage{14268}
(\byear{2023})
\end{barticle}
\endbibitem

\bibitem[\protect\citeauthoryear{Shetty}{2023}]{shetty2023poorvi}
\begin{bchapter}
\bauthor{\bsnm{Shetty}, \binits{P.}}:
\bctitle{Poorvi@dravidianlangtech: Sentiment analysis on code-mixed tulu and tamil corpus}.
In: \bbtitle{Proceedings of the Third Workshop on Speech and Language Technologies for Dravidian Languages},
pp. \bfpage{124}--\blpage{132}
(\byear{2023})
\end{bchapter}
\endbibitem

\bibitem[\protect\citeauthoryear{She and Jia}{2021}]{she2021bigru}
\begin{barticle}
\bauthor{\bsnm{She}, \binits{D.}},
\bauthor{\bsnm{Jia}, \binits{M.}}:
\batitle{A bigru method for remaining useful life prediction of machinery}.
\bjtitle{Measurement}
\bvolume{167},
\bfpage{108277}
(\byear{2021})
\end{barticle}
\endbibitem

\bibitem[\protect\citeauthoryear{Subramanian et~al.}{2022}]{subramanian2022offensive}
\begin{barticle}
\bauthor{\bsnm{Subramanian}, \binits{M.}},
\bauthor{\bsnm{Ponnusamy}, \binits{R.}},
\bauthor{\bsnm{Benhur}, \binits{S.}},
\bauthor{\bsnm{Shanmugavadivel}, \binits{K.}},
\bauthor{\bsnm{Ganesan}, \binits{A.}},
\bauthor{\bsnm{Ravi}, \binits{D.}},
\bauthor{\bsnm{Shanmugasundaram}, \binits{G.K.}},
\bauthor{\bsnm{Priyadharshini}, \binits{R.}},
\bauthor{\bsnm{Chakravarthi}, \binits{B.R.}}:
\batitle{Offensive language detection in tamil youtube comments by adapters and cross-domain knowledge transfer}.
\bjtitle{Computer Speech \& Language}
\bvolume{76},
\bfpage{101404}
(\byear{2022})
\end{barticle}
\endbibitem

\bibitem[\protect\citeauthoryear{Sreelakshmi et~al.}{2024}]{sreelakshmi2024}
\begin{barticle}
\bauthor{\bsnm{Sreelakshmi}, \binits{K.}},
\bauthor{\bsnm{Premjith}, \binits{B.}},
\bauthor{\bsnm{Chakravarthi}, \binits{B.R.}},
\bauthor{\bsnm{Soman}, \binits{K.P.}}:
\batitle{Detection of hate speech and offensive language codemix text in dravidian languages using cost-sensitive learning approach}.
\bjtitle{IEEE Access}
\bvolume{12},
\bfpage{20064}--\blpage{20090}
(\byear{2024})
\end{barticle}
\endbibitem

\bibitem[\protect\citeauthoryear{Wang et~al.}{2021}]{Wang2021}
\begin{botherref}
\oauthor{\bsnm{Wang}, \binits{H.}}, et al.:
A survey on deep learning approaches for text classification.
IEEE Transactions on Knowledge and Data Engineering
(2021)
\end{botherref}
\endbibitem

\bibitem[\protect\citeauthoryear{Yao et~al.}{2024}]{yao2024comparison}
\begin{bchapter}
\bauthor{\bsnm{Yao}, \binits{Q.}},
\bauthor{\bsnm{Kollmeyer}, \binits{P.J.}},
\bauthor{\bsnm{Lu}, \binits{D.D.-C.}},
\bauthor{\bsnm{Emadi}, \binits{A.}}:
\bctitle{A comparison study of unidirectional and bidirectional recurrent neural network for battery state of charge estimation}.
In: \bbtitle{2024 IEEE Transportation Electrification Conference and Expo (ITEC)},
pp. \bfpage{1}--\blpage{6}.
\bpublisher{IEEE}, \blocation{???}
(\byear{2024})
\end{bchapter}
\endbibitem

\bibitem[\protect\citeauthoryear{Zhao et~al.}{2018}]{zhao2018we}
\begin{botherref}
\oauthor{\bsnm{Zhao}, \binits{X.}},
\oauthor{\bsnm{Feng}, \binits{G.C.}},
\oauthor{\bsnm{Liu}, \binits{J.S.}},
\oauthor{\bsnm{Deng}, \binits{K.}}:
We agreed to measure agreement—redefining reliability de-justifies krippendorff's alpha.
China Media Research
\textbf{14}(2)
(2018)
\end{botherref}
\endbibitem

\bibitem[\protect\citeauthoryear{Zucchet and Orvieto}{2024}]{zucchet2024recurrent}
\begin{barticle}
\bauthor{\bsnm{Zucchet}, \binits{N.}},
\bauthor{\bsnm{Orvieto}, \binits{A.}}:
\batitle{Recurrent neural networks: vanishing and exploding gradients are not the end of the story}.
\bjtitle{Advances in Neural Information Processing Systems}
\bvolume{37},
\bfpage{139402}--\blpage{139443}
(\byear{2024})
\end{barticle}
\endbibitem

\bibitem[\protect\citeauthoryear{Zhang et~al.}{2018a}]{zhang2018deep}
\begin{barticle}
\bauthor{\bsnm{Zhang}, \binits{L.}},
\bauthor{\bsnm{Wang}, \binits{S.}},
\bauthor{\bsnm{Liu}, \binits{B.}}:
\batitle{Deep learning for sentiment analysis: A survey}.
\bjtitle{Wiley Interdisciplinary Reviews: Data Mining and Knowledge Discovery}
\bvolume{8}(\bissue{4}),
\bfpage{1253}
(\byear{2018})
\end{barticle}
\endbibitem

\bibitem[\protect\citeauthoryear{Zhang et~al.}{2018b}]{Zhang2018}
\begin{barticle}
\bauthor{\bsnm{Zhang}, \binits{L.}},
\bauthor{\bsnm{Wang}, \binits{S.}},
\bauthor{\bsnm{Liu}, \binits{B.}}:
\batitle{Deep learning for sentiment analysis: A survey}.
\bjtitle{Wiley Interdisciplinary Reviews: Data Mining and Knowledge Discovery}
\bvolume{8}(\bissue{4}),
\bfpage{1253}
(\byear{2018})
\end{barticle}
\endbibitem

\end{thebibliography}



\end{document}